\documentclass[10pt,twocolumn,letterpaper]{article}
\usepackage[table]{xcolor}
\usepackage{iccv}
\usepackage{times}
\usepackage{amsmath}
\usepackage{amssymb}
\usepackage{booktabs}
\usepackage{caption}
\usepackage{cuted}
\usepackage{graphicx}
\usepackage{pifont}
\usepackage{float}
\usepackage{pythonhighlight}
\usepackage[breaklinks=true,bookmarks=false,colorlinks]{hyperref}
\usepackage[capitalise]{cleveref}
\usepackage[square,numbers,sort]{natbib}
\usepackage{multirow}
\iccvfinalcopy
\def\iccvPaperID{9378}

\ificcvfinal\fi

\makeatletter
\renewcommand{\paragraph}{%
  \@startsection{paragraph}{4}%
  {\z@}{.5ex \@plus 1ex \@minus .2ex}{-1em}%
  {\normalfont\normalsize\bfseries}%
}
\makeatother

\makeatletter
\DeclareRobustCommand\onedot{\futurelet\@let@token\@onedot}
\def\@onedot{\ifx\@let@token.\else.\null\fi\xspace}

\makeatother

\begin{document}
\title{Do You Even Need Attention? A Stack of Feed-Forward Layers Does Surprisingly Well on ImageNet}

\author{Luke Melas-Kyriazi\\
Oxford University\\
{\tt\small lukemk@robots.ox.ac.uk}
}

\maketitle
\ificcvfinal\thispagestyle{plain}\fi

% \begin{strip}
% % \vspace{-3.0em}
% \begin{center}
% \fbox{\makebox(500,100){}}
% % \includegraphics[width=\linewidth]{images/Splash-Figure-v4.pdf}
% \captionof{figure}{Splash Figure Caption}
% \label{f:splash}
% \end{center}
% \end{strip}

\begin{abstract}
The strong performance of vision transformers on image classification and other vision tasks is often attributed to the design of their multi-head attention layers. However, the extent to which attention is responsible for this strong performance remains unclear. In this short report, we ask: is the attention layer even necessary? Specifically, we replace the attention layer in a vision transformer with a feed-forward layer applied over the patch dimension. The resulting architecture is simply a series of feed-forward layers applied over the patch and feature dimensions in an alternating fashion. In experiments on ImageNet, this architecture performs surprisingly well: a ViT/DeiT-base-sized model obtains 74.9\% top-1 accuracy, compared to 77.9\% and 79.9\% for ViT and DeiT respectively. These results indicate that aspects of vision transformers other than attention, such as the patch embedding, may be more responsible for their strong performance than previously thought. We hope these results prompt the community to spend more time trying to understand why our current models are as effective as they are.\footnote{Code is available at \href{https://github.com/lukemelas/do-you-even-need-attention}{https://github.com/lukemelas/do-you-even-need-attention}}
\end{abstract}

\section{Introduction}\label{s:intro}

Introduced by \cite{dosovitskiy2020image}, the vision transformer architecture applies a series of transformer blocks to a sequence of image patches. Each block consists of a multi-head attention layer~\cite{vaswani2017attention} followed by a feed-forward layer (i.e. a linear layer, or a single-layer MLP) applied along the feature dimension. The general-purpose nature of this architecture, coupled with its strong performance on image classification benchmarks, has prompted significant interest from the vision community. However, it is still not clear exactly \textit{why} the vision transformer is effective. 

The most-cited reason for the transformer's success on vision tasks is the design of its attention layer, which gives the model a global receptive field. This layer may be seen as a data-dependent linear layer, and when applied on image patches it resembles (but is not exactly equivalent to) a convolution. Indeed, a significant amount of recent work has gone into improving the efficiency and efficacy of the attention layer. 

\begin{figure}
  \centering
  \includegraphics[width=0.35\textwidth]{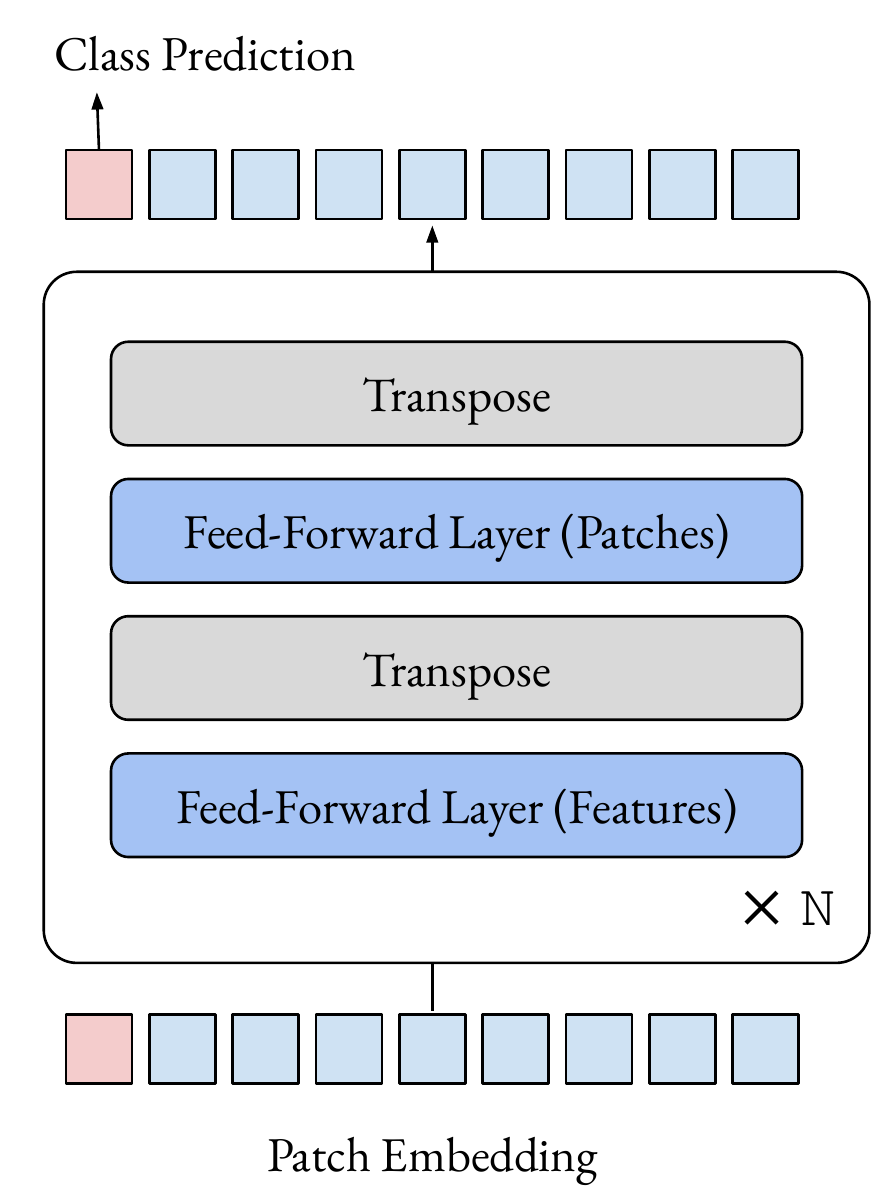}
  \caption{The architecture explored in this report is extremely simple, consisting of a patch embedding followed by a series of feed-forward layers. These feed-forward layers are alterately applied to the patch and feature dimensions of the image tokens. The architecture is identical to that of ViT~\cite{dosovitskiy2020image} with the attention layer replaced by a feed-forward layer.}\label{fig:comparison}
\end{figure}

In this short report, we conduct an experiment that hopes to shed a little light on why the vision transformer works so well in the first place. Specifically, we remove attention from the vision transformer, replacing it with a feed-forward layer applied over the patch dimension. After this change, the model is simply a series of feed-forward layers applied over the patch and feature dimensions in an alternating fashion (\autoref{fig:comparison}). 

In experiments on ImageNet (\autoref{table:comparison}), we show that quite strong performance is attainable even without attention. Notably, a ViT-base-sized model gives 74.9\% top-1 accuracy without any hyperparameter tuning (i.e. using the same hyperparameters as its ViT counterpart). These results suggest that the strong performance of vision transformers may be attributable less to the attention mechanism and more to other factors, such as the inducive bias produced by the patch embedding and the carefully-curated set of training augmentations. 

The primary purpose of this report is to explore the limits of simple architectures. We do not to break the ImageNet benchmarks; on that front, methods such as neural architecture search (e.g. EfficientNet~\cite{tan2019efficientnet}) will inevitably perform best. Nonetheless, we hope that the community finds these results interesting, and that these results prompt more researchers to investigate why our current models are as effective as they are.

\section{Background}\label{s:background}

The context for this report is that over the past few months, there has been an explosion of research into variants of the vision transformer architecture: Deit~\cite{touvron2020deit} adds distillation, DeepViT~\cite{zhou2021deepvit} mixes the attention heads, CaiT~\cite{touvron2021cait} separates the attention layers into two stages, Token-to-Token ViT~\cite{yuan2021tokens} aggregates neighboring tokens throughout the network, CrossViT~\cite{chen2021crossvit} processes patches at two scales, PiT~\cite{heo2021rethinking} adds pooling layers, LeViT~\cite{graham2021levit} uses convolutional embeddings and modified attention/normalization layers, CvT~\cite{wu2021cvt} uses depthwise convolutions in the attention layer, and Swin/Twins~\cite{liu2021swin,chu2021twins} combines global and local attention, just to name a few. 

These works improve upon the vision transformer architecture, each showing strong performance on ImageNet. However, it is not clear how the different parts of ViT or its many variants contribute to the final performance of each of these models. This report details an experiment that investigates one aspect of this issue, namely how important the attention layers are to ViT's success. 

\section{Method and Experiments}\label{s:methods}

\subsection{Do You Even Need Attention?}

We remove the attention layer from the ViT model, replacing it by a simple feed-forward layer over the patch dimension. We use the same structure as the standard feed-forward network over the feature dimension, which is to say that we project the patch dimension into a higher-dimensional space, apply a nonlinearity, and project it back to the original space. \autoref{fig:code} gives PyTorch code for a single block of the feed-forward-only transformer.

We should note that, like the vision transformer and its many variants, this feed-forward-only network bares strong resemblance to a convolutional network. In fact, the feed-forward layer over the patch dimension can be viewed as an unusual type of convolution with a full receptive field and a single channel. Since the feed-forward layer over the feature dimension can be seen as a 1x1 convolution, it would be technically accurate to say that the entire network is a sort of convolutional network in disguise. That being said, it is structurally more similar to a transformer than to a traditionally-designed convolutional network (e.g. ResNet/VGG). 

\begin{table}[t]
  \small
  \centering
  % \def\arraystretch{1.15}
  % \rowcolors{2}{gray!10}{white}
  \begin{tabular}{|c|l|c|c|} \toprule
                                &         & Params & ImageNet Top-1 \\ \midrule
\multirow{3}{*}{Tiny $(P=16)$}  & ViT     & -               & -              \\
                                & DeiT    & 5.7M            & 72.2           \\
                                & FF Only & 7.7M            & 61.4           \\ \midrule
\multirow{3}{*}{Base $(P=16)$}  & ViT     & 86M             & 77.9           \\
                                & DeiT    & 86M             & 79.9           \\
                                & FF Only & 62M             & 74.9           \\ \midrule
\multirow{3}{*}{Large $(P=32)$} & ViT     & 306M            & 71.2           \\
                                & DeiT    & -               & -              \\
                                & FF Only & 206M            & 71.4          \\ \bottomrule
\end{tabular}
  \caption{A comparison of ImageNet top-1 accuracies for different model sizes. In the first column, $P$ refers to the patch size in pixels. Overall, the models with only feed-forward layers (\textit{FF Only}) perform worse than their counterparts with attention, but they perform surprisingly well regardless. Performance deteriorates for the largest models both with and without attention.}
  \label{table:comparison}
  \end{table}

\subsection{Experimental Setup}

We train three models, corresponding to the ViT/DeiT tiny, base, and large networks, on ImageNet~\cite{imagenet_dataset} using the setup from DeiT~\cite{touvron2020deit}. The tiny and base networks have patch size 16, while the large network has patch size 32 due to computational constraints. Training and evaluation is performed at resolution 224px. Notably, we use exactly same hyperparameters as DeiT for all models, which means that our performance could likely be improved with hyperparameter tuning. 

\subsection{Results}

\autoref{table:comparison} shows the performance of our simple feed-forward network on ImageNet. Most notable, the feed-forward-only version of ViT/Deit-base achieves surprisingly strong performance (74.9\% top-1 accuracy), comparable to a number of older convolutional networks (e.g. VGG-16, ResNet-34). Such a comparison is not exactly fair because the feed-forward model uses stronger training augmentations, but it is nevertheless quite a strong result in an absolute sense. 

Performance deteriorates for the large models for the models both with and without attention, yeilding 71.2\% and 71.4\% top-1 accuracy respectively. As detailed in \cite{dosovitskiy2020image}, pretraining with a larger dataset appears necessary for such enormous models. 

\subsection{Do You Even Need Feed-Forward Layers?}

Naturally, since we tried training a model with only feed-forward layers, we also tried training a model with only attention layers. In this model, we simply replaced the feed-forward layer over the feature dimension with an attention layer over the feature dimension. We only experimented with a tiny-sized model (~4.0M parameters), but it performed spectacularly poorly (28.2\% top-1 accuracy at 100 epochs, at which point we ended the run). 

\subsection{Discussion}

The above experiments demonstrate that it is possible to train a reasonably strong transformer-style image classifiers without attention layers. Furthermore, attention layers without feed-forward layers do not appear to yield similarly strong performance. These results indicate that the strong performance of ViT may be attributable more to its patch embeddings and training procedure than to the design of the attention layer. The patch embeddings in particular provide a strong inductive bias that is likely one of, if not the, main drivers of the model's strong performance. 

% As mentioned above, the purpose of this report is to better understand vision transformers, not to develop new state-of-the-art architectures. 

From a practical perspective, the feed-forward-only model has one notable advantage over the vision transformer, which is that its complexity is linear with respect to the sequence length as opposed to quadratic. This is the case due to the intermediate projection dimension within the feed-forward layer applied over patches, whose size is not necessarily dependent on the sequence length. Usually the intermediate dimension is chosen to be a multiple of the number of input features (i.e. the number of patches), in which case it the model is indeed quadratic, but this does not necessarily need to the case. 

Apart from its worse performance, one major downside of the feed-forward-only model is that it only functions on fixed-length sequences (due to the feed-forward layer over patches). This is not a big issue for image classificaion, where images are cropped to a standard size, but limits the applicability of the architecture to other tasks. 

Feed-forward-only models shed light on vision transformers and attention mechanisms in general. For the future, it would be interesting to investigate the extent to which these conclusions apply outside of the image domain, for example in NLP/audio. 

\subsection{Conclusion}

This short report demonstrates that transformer-style networks without attention layers make for surprisingly strong image classifiers. Future work in this direction could attempt to better understand the contributions of other pieces of the transformer architecture (e.g. the normalization layer or initialization scheme). More broadly, we hope that this short report encourages further investigation into why our current models perform as well as they do. 

\begin{figure*}
  \centering
  \inputpython{figures/code.py}{0}{100}
  \caption{PyTorch code for a single transformer block consisting of two feed-forward layers}\label{fig:code}
\end{figure*}

{\small\bibliographystyle{ieee_fullname}\bibliography{bibliography/new,bibliography/vgg_bibtex/shortstrings,bibliography/vgg_bibtex/longstrings,bibliography/vgg_bibtex/vgg_local,bibliography/vgg_bibtex/vgg_other}}

% \clearpage
% \input{6-appendix}

\end{document}